\documentclass[runningheads]{llncs}
\usepackage{graphicx}
\usepackage{hyperref}
\let\svthefootnote\thefootnote
\newcommand\freefootnote[1]{%
  \let\thefootnote\relax%
  \footnotetext{#1}%
  \let\thefootnote\svthefootnote%
}
\usepackage[T1]{fontenc}
\usepackage[utf8]{inputenc} 
\usepackage{amssymb,amsmath,array}
\usepackage[table,svgnames]{xcolor}
\usepackage{algorithm}
\usepackage{algpseudocode}
\usepackage{multirow}
\usepackage[section]{placeins}
\usepackage{orcidlink}
\begin{document}

%
%\title{Contribution Title\thanks{Supported by organization x.}}
\title{Logic-based Explanations for Linear Support Vector Classifiers with Reject Option}
%\thanks{Partially supported by FUNCAP.}
%
\titlerunning{Logic-based Explanations for Linear SVC with Reject Option}
 \author{Francisco Mateus {Rocha Filho}\orcidlink{0009-0002-2261-3767} \and
 Thiago Alves {Rocha}\orcidlink{0000-0001-7037-9683} \and
 Reginaldo Pereira {Fernandes Ribeiro}\orcidlink{0009-0000-0250-5713} \and
 Ajalmar Rêgo da {Rocha Neto}\orcidlink{0000-0002-4512-5531}}
% \author{Francisco Mateus Rocha~Filho\inst{1}\orcidID{0000-1111-2222-3333} \and
% Reginaldo Pereira Fernandes~Ribeiro\inst{2,3}\orcidID{1111-2222-3333-4444} \and
% Thiago Alves Rocha\inst{3}\orcidID{2222--3333-4444-5555} \and
% Ajalmar Rêgo da Rocha~Neto \inst{3}\orcidID{2222--3333-4444-5555}}
%
\authorrunning{F. M. Rocha Filho et al.}
\institute{
Instituto Federal de Educação, Ciência e Tecnologia do Ceará (IFCE), Brazil
\email{\{francisco.mateus.rocha06\}@aluno.ifce.edu.br}\\
\email{\{thiago.alves,reginaldo.fernandes,ajalmar\}@ifce.edu.br}
}
\maketitle              % typeset the header of the contribution
\begin{abstract}
Support Vector Classifier (SVC) is a well-known Machine Learning (ML) model for linear classification problems. It can be used in conjunction with a reject option strategy to reject instances that are hard to correctly classify and delegate them to a specialist. This further increases the confidence of the model. Given this, obtaining an explanation of the cause of rejection is important to not blindly trust the obtained results. While most of the related work has developed means to give such explanations for machine learning models, to the best of our knowledge none have done so for when reject option is present. We propose a logic-based approach with formal guarantees on the correctness and minimality of explanations for linear SVCs with reject option. We evaluate our approach by comparing it to Anchors, which is a heuristic algorithm for generating explanations. Obtained results show that our proposed method gives shorter explanations with reduced time cost.
%a specialist can use the explanation for a rejection to correctly classify the instance
%We evaluate our approach by comparing it to Anchors, which is a heuristic algorithm for generating explanations.

\end{abstract}
%Linear programming \and
%
%
%
\section{Introduction}\label{Introduction}
%\textcolor{red}{Links de instrução para o artigo: https://resource-cms.springernature.com/springer-cms/rest/v1/content/19242230/data/v11\\ %llncsdoc.pdf included in this manuscript, left side menu}

\freefootnote{\tiny{This preprint has
not undergone peer review or any post-submission improvements or
corrections. The Version of Record of this contribution is published in Intelligent Systems, LNCS, vol 14195, and is available online at \href{https://doi.org/10.1007/978-3-031-45368-7_10}{https://doi.org/10.1007/978-3-031-45368-7\underline{ }10}.}}
%DIZER QUE NOSSA ABORDAGEM PODE SER FACILMENTE ADAPTADA PARA OUTROS MODELOS DE ML COM OPCAO DE REJEICAO

It is undeniable that Artificial Intelligence is increasingly being inserted into the daily lives of people~\cite{lecun2015deep}, influencing the most complex decision taking tasks~\cite{Daniel2018Intelligible}. Consequently, the most varied classification models in machine learning may come across instances that are difficult to correctly classify, be it due to lack of good data, to feature bias (color, size, gender)~\cite{ignatiev2020towards} or to noise present on the given patterns~\cite{Daniel2018Intelligible}.

The SVC is one of the most well-known ML models. The linear Support Vector Machine (SVM), specifically, has been used in a variety of classification problems~\cite{chlaoua2019deep,richhariya2020diagnosis,yang2018automatic}. However, the model can also fall into the same previously said pitfalls, failing to give a correct classification. As such, the reject option strategy depicted in~\cite{chow} can be used to remedy such cases. Classification with reject option (RO) is a paradigm that aims to improve reliability in decision support systems by rejecting the classification results of more complex cases and avoiding a higher error rate. These cases are separated to be dealt with in another specialized way, be it through other models or with the assistance of a human specialist~\cite{mesquita2016classification}. Since classification with reject option comprises a set of techniques, in this work we consider the strategy where the classifier is trained as usual and the rejected cases are determined after the training phase. Usually, this process requires finding a trade-off between the costs of misclassifications and
rejections.

%\textcolor{red}{ELABORAR UM POUCO MAIS SOBRE REJECT OPTION. Mateus: aceito sugestões do que colocar aqui. Talvez sobre o empirical risk? Ou que já tem exemplos aplicados em SVM.}
%As such, the reject option strategy depicted in~\cite{chow} can be used to remedy such cases, separating the more complex cases to be dealt with in another, specialized way, be it through other models or a human~\cite{mesquita2016classification}.

The linear SVC can already be globally interpreted to a certain extent based on the decision function, where the most important features are often associated with the highest weights \cite{guyon2002gene}. Such analysis, however, is not capable of giving decisive answers on more specific cases \cite{krause2017workflow}. An example of this is depicted and more thoroughly explored in Section \ref{Linear_SVCs_May_Not_Be_Locally_Interpretable}. This gives a margin for questionable explanations, especially for when reject option is present due to the added level of complexity. Therefore the need for a more thorough, instance-based explanation method.

Then, in this work, we consider instance-based explanations. Specifically, the objective of instance-based explanations is to provide interpretable insights by highlighting relevant features that influenced the output of an ML model on a given instance. By linking the output to specific instances and their features, users can gain a better understanding of the reasons behind the predictions of ML models.

%It is when the computation of an explanation rigorously explores the entire feature space of a given instance. Due to this, such explanations are provably correct and hold for any point in the space, which therefore makes them trustworthy~\cite{ignatiev2020towards}.

The popularization of concepts of Explainable Artificial Intelligence (XAI) increased efforts to explain most complex models~\cite{ignatiev2020onFormal}. These have mostly been done through heuristic methods, the more prevalent approach, such as LIME~\cite{Ribeiro2016Lime}, SHAP~\cite{Lundberg2017Shap}, and Anchors~\cite{Ribeiro2018Anchors}. These tend to be model-agnostic and not able to guarantee a trustworthy explanation with regard to the characteristics of the model or the correctness of the answers~\cite{ignatiev19onValidating}. Then, their explanations can be often proven wrong through counterexamples that expose their contradictions~\cite{ignatiev19onValidating,ignatiev2020onFormal}, leading to even more doubts regarding how much the ML model can be trusted.

% \textcolor{red}{adicionar citação. Ignatiev: Abduction-Based Explanations for Machine Learning Models. Pode adicionar outras citações. Do jeito que tá, parece que ninguém fez isso antes.}

Thus, the importance of trustworthy ML models increases the need for logic-based approaches to explain the decisions made by these models~\cite{ignatiev2019abduction,ignatiev2020onFormal}. The computation of explanations in these approaches rigorously explores the entire feature space. Due to this, such explanations are provably correct and hold for any point in the space, which therefore makes them trustworthy~\cite{ignatiev2020towards}. Moreover, logic-based approaches can guarantee minimality of explanations. Minimality is important since succinct explanations seem easier to be interpreted by humans.
Recent years have seen a surge in research dedicated to investigating logic-based XAI for ML models, such as neural networks, naive Bayes, random forests, decision trees, and boosted trees~\cite{audemard2022preferred,ignatiev2022using,ignatiev2019abduction,izza2020explaining,ijcai2021p0366,marques2020explaining}.

Due to the importance of reject option and logic-based explainability for trustworthy ML models, this work proposes a logic-based approach to explain linear SVCs with reject option. Given that rejected instances may be further analyzed by specialists, explanations regarding the causes of rejections can reduce the workload of these professionals. Our proposal builds on work from the literature of logic-based explainability for traditional ML models, i.e. without reject option~\cite{ignatiev2019abduction,ignatiev19onValidating}. We compute explanations for linear SVCs with reject option by solving a set of logical constraints, specifically, boolean combinations of linear constraints. Despite the fact that we consider a linear SVC with reject option in this work, our approach can be easily adapted for other ML models with reject option, such as neural networks.
%, as neural networks, decision trees, random forests, and boosted trees
%IMPORTE PARA EXPLICAR OS MOTIVOS PELA REJEICAO. PODE AJUDAR O ESPECIALISTA A TOMAR A DECISAO!
%Neural networks for soft decision making
% For this method, the classifier is trained as usual (i.e., without
% referring to an explicit rejection class); but rather, the rejection
% region is determined after the training phase, heuristically or
% based on the optimization of some post-training criterion that
% weighs the trade-off between the costs of misclassification and
% rejection.

%This paper uses a linear SVC with reject option as the target for generating correct explanations. The decision function is used for transforming the classification problem into a logic-based problem for instance-based explanations. 

We conducted experiments to compare our approach against Anchors, a heuristic method that generates explanations by locally exploring the feature subspace close to a given instance~\cite{Ribeiro2018Anchors}, through six different datasets. The results show that our approach is capable to generate succinct explanations up to 286 times faster than Anchors, in scenarios with and without the presence of reject option.

\section{Background}

\subsection{Machine Learning and Binary Classification Problems}\label{Machine_Learning_and_Binary_Classification_Problems}

In machine learning, binary classification problems are defined over a set of features $\mathcal{F} = \{f_1, ..., f_n\}$ and a set of two classes $\mathcal{K} = \{-1, +1\}$. In this paper, we consider that each feature $f_i \in \mathcal{F}$ takes its values $x_i$ from the domain of real numbers. Moreover, each feature $f_i$ has an upper bound $u_i$ and a lower bound $l_i$ such that $l_i \leq x_i \leq u_i$. Then, each feature $f_i$ has domain $[l_i, u_i]$. Besides, the notation $\mathbf{x} = \{f_1 = x_1, f_2 = x_2, ..., f_n = x_n\}$ represents a specific point or instance such that each $x_i$ is in the domain of $f_i$.%\in {\rm I\!R}$.
%the values $x_i$ of $f_i$ are defined by

A binary classifier $C$ is a function that maps elements in the feature space into the set of classes $\mathcal{K}$. For example, $C$ can map instance $\{f_1 = x_1, f_2 = x_2, ..., f_n = x_n\}$ to class $+1$. Usually, the classifier is obtained by a training process given as input a training set $\{ \mathbf{x}_i,y_i\}^{l}_{i=1}$, where $\mathbf{x}_i \in {\rm I\!R}^n$ is an input vector or pattern and $y_i \in \{-1, +1\}$ is the respective class label. Then, for each input vector $\mathbf{x}_i$, its input values $x_{i, 1}$, $x_{i, 2}$, ..., $x_{i, n}$ are in the domain of corresponding features $f_1$, ..., $f_n$. A well-known classifier and its training process are presented in Subsection~\ref{Support_Vector_Machine}.
%the next subsection.

\subsection{Support Vector Machine}\label{Support_Vector_Machine}
The SVM~\cite{cortes1995support} is a supervised machine learning model often used for classification problems. It uses the concept of an optimal separating hyperplane depicted in~\cite{boser1992training,cortes1995support} to separate the data. On a ${\rm I\!R}^n$ space, such a hyperplane is defined by a set of points $\mathbf{x}$ that satisfies 
\begin{equation}\label{hyperplane_eq}
    \mathbf{w}_o \cdot \mathbf{x} + b = 0,
\end{equation}
where $\mathbf{w}_o \in {\rm I\!R}^n$ is the optimal weight vector, $\mathbf{x} \in {\rm I\!R}^n $ is a feature vector with $n$ features and an intercept (bias) $b \in {\rm I\!R}$.

%, where $x_i \in {\rm I\!R}^n$ is an input vector and $y_i \in \{-1, +1\}$ is the respective class label, 

A given training set $\{ \mathbf{x}_i,y_i\}^{l}_{i=1}$ is said to be linearly separable if there is $\mathbf{w}_o \in {\rm I\!R}^n$ and $b \in {\rm I\!R}$ that guarantees the separation between positive and negative class patterns without error. In other words, the following inequalities
\begin{equation}
\text{for }i \in \{1, ..., l\},
    \begin{cases}
        \begin{aligned}
         \mathbf{w}_o \cdot \mathbf{x}_i + b \geq +1,\quad & \text{if } y_i = +1,\\
         \mathbf{w}_o \cdot \mathbf{x}_i + b \leq -1,\quad & \text{if } y_i = -1,   
        \end{aligned}    
    \end{cases}
\end{equation}
must be satisfied to obtain the optimal hyperplane $h_o = \{ \mathbf{x} \mid \mathbf{w}_o \cdot \mathbf{x} + b = 0\}$.

A Hard Margin SVM (SVM-HM)~\cite{Vapnik1998Statistical} can be used when the data is linearly separable and misclassifications are not allowed, maximizing the margin between two hyperplanes, $h_+$ and $h_-$, parallel to $h_o$. There can be no training patterns between $h_+$ and $h_-$. Once maximizing the margin is similar to  minimizing $\frac{1}{2}||\mathbf{w}||^2$, the optimization problem for obtaining the optimal parameters for the SVM can be described as follows:
\begin{equation}
\begin{aligned}
\min_{\mathbf{w},b} & \quad{\frac{1}{2}||\mathbf{w}||^2} \\
\textrm{s.t.} & \quad y_i(\mathbf{w} \cdot \mathbf{x}_i + b) \geq 1,\quad \text{for }i \in \{ 1,...,l\}.
\end{aligned}
\end{equation}
Thus, the decision function used for classifying input instances $\mathbf{x}$ is defined in the following:
\begin{equation}\label{decision_function}
    \begin{aligned}
        d(\mathbf{x}) = \mathbf{w} \cdot \mathbf{x} + b,
    \end{aligned}
\end{equation}
while the predicted label $\hat{y} \in \mathcal{K}$ of an input $\mathbf{x}$ is given by
\begin{equation}\label{predicition_output}
    \hat{y}= % p(\mathbf{x}) = 
    \begin{cases}
        \begin{aligned}
         +1, \quad \text{if } d(\mathbf{x}) > 0,\\
         -1, \quad \text{if } d(\mathbf{x}) < 0. 
        \end{aligned}    
    \end{cases}
\end{equation}

However, in real-world problems, the training patterns from the two classes can not be linearly separated by a hyperplane due to data overlapping. In order to overcome this situation, one must use Soft-Margin SVMs (SVM-SM) in which misclassifications are allowed to happen. To do so, slack variables can be used to relax the constraints of the SVM-HM~\cite{cortes1995support,Vapnik1998Statistical}. Given this, the optimization problem for Soft-Margin SVMs is described as
\begin{equation}
\begin{aligned}
\min_{\mathbf{w},b,\mathbf{\xi}} & \quad \frac{1}{2}||\mathbf{w}||^2 - C\sum^{l}_{i=1}\xi_i \\
\textrm{s.t.} & \quad y_i(\mathbf{w} \cdot \mathbf{x}_i + b) \geq 1 - \xi_i,\quad & \text{for } i \in \{1,...,l\}\\
                    & \quad \xi_i \geq 0,\quad & \text{for } i \in \{1,...,l\}
\end{aligned}
\end{equation}
where $C$ is a trade-off between $\frac{1}{2}||\mathbf{w}||^2$ and $\sum_{i}^n \xi_i$. Thus, for a high enough value of $C$, minimizing the sum of errors while maximizing the separation margin leads toward the optimal hyperplane. This can be the same as the one found through SVM-HM if the data is linearly separable. Moreover, the decision function and predicted label for SVM-HM and SVM-SM are the same.
% The Lagrangian function can be defined as

% \begin{equation}
% \begin{aligned}
%  &({\frac{1}{2}||w||^2 + C\sum^{l}_{i=1}\xi_i - \sum^{l}_{i=1}\alpha_i(y_i(w^Tx_i + b) -1 + \xi_i )} 
%  -\sum^{l}_{i=1}\beta_i\xi_i).\\
% \end{aligned}
% \end{equation}
% where $\alpha_{i=1}^{n}$, $\beta_{i=1}^{n}$ and $\xi_{i=1}^{n}$ all have non-negative values.

% Minimizing this through the primal problem variables enables and maximizing through the dual problem variables gives space to the optimization problem
% \begin{equation}
% \begin{aligned}
% \max L(\alpha) = \quad & \sum^{l}_{i=1}\alpha_i - \frac{1}{2}\sum^{l}_{i=1}\sum^{l}_{j=1}\alpha_i\alpha_j y_i y_j(x_i{^T} x_j)\\
% \textrm{s.t.} \quad & \sum^{l}_{i=1}\alpha_i y_i = 0\\
% &0 \leq \alpha_i \leq C, \quad i = 1,...,l,
% \end{aligned}
% \end{equation}
% which in turn leads to the optimal solution.

\subsection{Reject Option Classification}\label{Reject_Option_Classification}
Reject option for classification problems, as depicted in~\cite{chow}, is a set of techniques that aim to improve reliability in decision support systems. In the case of a binary problem, it consists in withholding and rejecting  a classification result that is ambiguous enough, i.e. when the instance is too close to the decision boundary of the classifier. For the linear SVC, it is when the instance is too close to the separating hyperplane. Then, these rejected instances are analyzed through another classification method or even by a human specialist~\cite{mesquita2016classification}.

In applications where a high degree of reliability is needed and misclassifications can be too costly, rejecting to classify a pattern can be more beneficial than the risk of a higher error rate due to wrong classifications~\cite{DeOliveira2016efficient}. According to~\cite{chow}, the optimal classifiers that best handle such a relation can be achieved by the minimization of the empirical risk
\begin{equation}
    \begin{aligned}
        \hat{R} = E + w_r R
    \end{aligned}
\end{equation}
%TODO: Olhar notacao do w_r
where $R$ is the ratio of the number of rejected training patterns to the number of patterns in the entire training dataset; $E$ is the ratio of the number of misclassified patterns to the number of all the training patterns without including those ones rejected; and $w_r$ is a weight denoting the cost of rejection. A lower $w_r$ gives room for a decreasing error rate at the cost of a higher quantity of rejected instances, with the opposite happening for a higher $w_r$.

A method is presented in~\cite{mesquita2016classification} for single, standard binary classifiers that do not provide probabilistic outputs. For SVCs, the proposed rejection techniques are based on the distance of patterns to the optimal separating hyperplane. If the distance value is lower than a predefined threshold, then the pattern is rejected. As such, a rejection region is determined after the training step of the classifier, with a threshold containing appropriate values being applied to the output of the classifier. Therefore, applying this strategy to a standard binary SVC leads to the following prediction cases:
\begin{equation}\label{reject_predicition_output}
    \hat{y} = 
    \begin{cases}
        \begin{aligned}
         &+1, \quad \text{if } f(\mathbf{x}) > t_+,\\
         &-1, \quad \text{if } f(\mathbf{x}) < t_-,\\
         &\hspace{0.45cm} 0, \quad \text{otherwise},
        \end{aligned}    
    \end{cases} 
\end{equation}
where $t_+$ and $t_-$ are the thresholds for the positive class and negative class, respectively, and 0 is the rejection class. Furthermore, these thresholds are chosen to generate the optimal reject region, which corresponds to the region that minimizes the empirical risk by producing both the ratio of misclassified patterns $E$ and the ratio of rejected patterns $R$.

%\textcolor{red}{Furthermore, the values for these thresholds are the ones that generate the best reject region, i.e. that outputs $E$ and $R$ that minimizes the empirical risk $\hat{R}$.}

%\textcolor{red}{DIZER EM POUCAS PALAVRAS COMO OS ts PODEM SER DEFINIDOS (DE ACORDO COM A MINIMIZACAO DO Rchapeu)}

\subsection{Heuristic-Based XAI}\label{Heuristic_based XAI}
Some ML models are able to be interpreted by nature, such as decision trees \cite{Ribeiro2016Lime}. Others, such as neural networks and boosted trees, have harder-to-explain outputs, leading to the use of specific methods to get some degree of explanation~\cite{ignatiev19onValidating}. One of the most predominant ways to achieve this is through the use of heuristic methods for generating instance-dependent explanations, which can be defined as a local approach. These analyze and explore the sub-space close to the given instance~\cite{ignatiev2020towards,ignatiev19onValidating}.
%A third type of model provides some level of interpretability, with artifices that enable explanations to some extent, i.e. linear SVMs.

Some of the most well-known heuristic methods are LIME, SHAP, and Anchors. These approaches are model-agnostic, generating local explanations while not taking into account the instance space as a whole \cite{Lundberg2017Shap,Ribeiro2018Anchors,Ribeiro2016Lime}. This, in turn, allows the explanations to fail when applied, since they can be consistent with instances predicted in different classes. Moreover, they can include irrelevant elements which could be otherwise removed while still maintaining the correctness of the answer~\cite{ignatiev2020onFormal}. Explanations with irrelevant elements may be harder to understand.

%not guaranteeing minimality and
%\cite{ignatiev2020towards}.
%giving a different prediction output than the one returned by the method. 

Anchors have been shown as a superior version to LIME, having a better accuracy with the resultant explanations~\cite{Ribeiro2018Anchors}. This rule-based method is designed to highlight which parts (features) of a given instance are sufficient for a classifier to make a certain prediction while being intuitive and easy to understand. However, it still gives room for wrong explanations due to the local characteristic, which can lead to cases where, for the same set of rules given by the explanation, different classes are predicted~\cite{ignatiev19onValidating}. Therefore, both explanation validity and size can be set as of questionable utility if they can not be fully relied upon.

%\subsection{Linear SVM feature weights may not be locally interpretable [subsection]}

\subsection{First-Order Logic}\label{First_Order_Logic}
%Logic-based Minimal Explanations

%over the function symbols and predicate symbols in $\{+, -, \times, =, \leq, <,  \geq, >\}$ such that each of these symbols has the standard meaning. 

In order to give explanations with guarantees of correctness, we use first-order logic (FOL)~\cite{kroening2016decision}. We use quantifier-free first-order formulas over the theory of linear real arithmetic. Then, first-order variables are allowed to take values from the real numbers. Therefore, we consider formulas as defined below:

\begin{equation}
        \begin{aligned}
             \varphi, \psi &:= s \mid (\varphi \wedge \psi) \mid (\varphi \vee \psi) \mid (\neg \varphi) \mid (\varphi \to \psi),
        \end{aligned}    
\end{equation}

\begin{equation}
        \begin{aligned}
             s &:= \sum^n_{i=1} a_i z_i \leq c \mid \sum^n_{i=1} a_i z_i < c,
        \end{aligned}    
\end{equation}
such that $\varphi$ and $\psi$ are quantifier-free first-order formulas over the theory of linear real arithmetic. Moreover, $s$ represents the atomic formulas such that $n \geq 1$, each $a_i$ and $c$ are concrete real numbers, and each $z_i$ is a first-order variable. For example, $(2.5z_1 + 3.1z_2 \geq 6) \wedge (z_1=1 \vee z_1=2) \wedge (z_1=2 \to z_2 \leq 1.1)$ is a formula by this definition. Observe that we allow standard abbreviations as $\neg (2.5z_1 + 3.1z_2 < 6)$ for $2.5z_1 + 3.1z_2 \geq 6$.

Since we are assuming the semantics of formulas over the domain of real numbers, an assignment $\mathcal{A}$ for a formula $\varphi$ is a mapping from the first-order variables of $\varphi$ to elements in the domain of real numbers. For instance, $\{z_1 \mapsto 2.3, z_2 \mapsto 1\}$ is an assignment for $(2.5z_1 + 3.1z_2 \geq 6) \wedge (z_1=1 \vee z_1=2) \wedge (z_1=2 \to z_2 \leq 1.1)$. An assignment $\mathcal{A}$ satisfies a formula $\varphi$ if $\varphi$ is true under this assignment. For example, $\{z_1 \mapsto 2, z_2 \mapsto 1.05\}$ satisfies the formula in the above example, whereas $\{z_1 \mapsto 2.3, z_2 \mapsto 1\}$ does not satisfy it. 

%To give an example, t
A formula $\varphi$ is satisfiable if there exists a satisfying assignment for $\varphi$. Then, the formula in the above example is satisfiable since $\{z_1 \mapsto 2, z_2 \mapsto 1.05\}$ satisfies it. As another example, the formula $(z_1 \geq 2) \wedge (z_1 < 1)$ is unsatisfiable since no assignment satisfies it. The notion of satisfiability can be extended to sets of formulas $\Gamma$. A set of first-order formulas is satisfiable if there exists an assignment of values to the variables that makes all the formulas in $\Gamma$ true simultaneously. For example, $\{(2.5z_1 + 3.1z_2 \geq 6), (z_1=1 \vee z_1=2), (z_1=2 \to z_2 \leq 1.1)\}$ is satisfiable given that $\{z_1 \mapsto 2, z_2 \mapsto 1.05\}$ jointly satisfies each one of the formulas in the set. It is well known that, for all sets of formulas $\Gamma$ and all formulas $\varphi$ and $\psi$,

\begin{equation}\label{or_unsat}
        \begin{aligned}
             \Gamma \cup \{\varphi \vee \psi\} \text{ is unsatisfiable iff } & \Gamma \cup \{\varphi\} \text{ is unsatisfiable and }\\
             & \Gamma \cup \{\psi\} \text{ is unsatisfiable.}
        \end{aligned}  
\end{equation}

Given a set $\Gamma$ of formulas and a formula $\varphi$, the notation $\Gamma \models \varphi$ is used to denote logical consequence, i.e., each assignment that satisfies all formulas in $\Gamma$ also satisfies $\varphi$. As an illustrative example, let $\Gamma$ be $\{z_1 = 2, z_2 \geq 1\}$ and $\varphi$ be $(2.5z_1 + z_2 \geq 5) \wedge (z_1=1 \vee z_1=2)$. Then, $\Gamma \models \varphi$ since each satisfying assignment for all formulas in $\Gamma$ is also a satisfying assignment for $\varphi$. Moreover, it is widely known that, for all sets of formulas $\Gamma$ and all formulas $\varphi$,

\begin{equation}\label{entailment_and_unsat}
        \begin{aligned}
             \Gamma \models \varphi \text{ iff } \Gamma \cup \{\neg \varphi\} \text{ is unsatisfiable.} 
        \end{aligned}  
\end{equation}

For instance, $\{ z_1 = 2, z_2 \geq 1, \neg((2.5z_1 + z_2 \geq 5) \wedge (z_1=1 \vee z_1=2)) \}$ has no satisfying assignment since an assignment that satisfies $(z_1 = 2 \wedge z_2 \geq 1)$ also satisfies $(2.5z_1 + z_2 \geq 5) \wedge (z_1=1 \vee z_1=2)$ and, therefore, does not satisfy $\neg((2.5z_1 + z_2 \geq 5) \wedge (z_1=1 \vee z_1=2))$.

Finally, we say that two first-order formulas $\varphi$ and $\psi$ are equivalent if, for each assignment $\mathcal{A}$, both $\varphi$ and $\psi$ are true under $\mathcal{A}$ or both are false under $\mathcal{A}$. We use the notation $\varphi \equiv \psi$ to represent that $\varphi$ and $\psi$ are equivalent. For example, $\neg ((z_1 + z_2 \leq 2) \wedge z_1 \geq 1)$ is equivalent to $(\neg(z_1 + z_2 \leq 2) \vee \neg (z_1 \geq 1))$. Besides, these formulas are equivalent to $((z_1 + z_2 > 2) \wedge z_1 < 1)$.

%TODO: satisfiability for a set of first-order formulas. DONE!
%TODO: equivalence of two formulas. DONE!

\section{Linear SVCs may not be Instance-based Interpretable}\label{Linear_SVCs_May_Not_Be_Locally_Interpretable}
%May Not Be Locally

Efforts have been made to bring explanations to linear SVCs prediction outputs. These have often been done through the analysis of the weights that compose the decision function, where the most important features are associated with the highest weights~\cite{guyon2002gene}. However, this may not be enough to enable correct interpretations. Although feature weights can give a rough indication of the overall relevance of features, they do not offer insight into the local decision-making process for a specific set of instances~\cite{krause2017workflow}.

\begin{figure}[!htpb]
\centering
\includegraphics[scale=0.7]{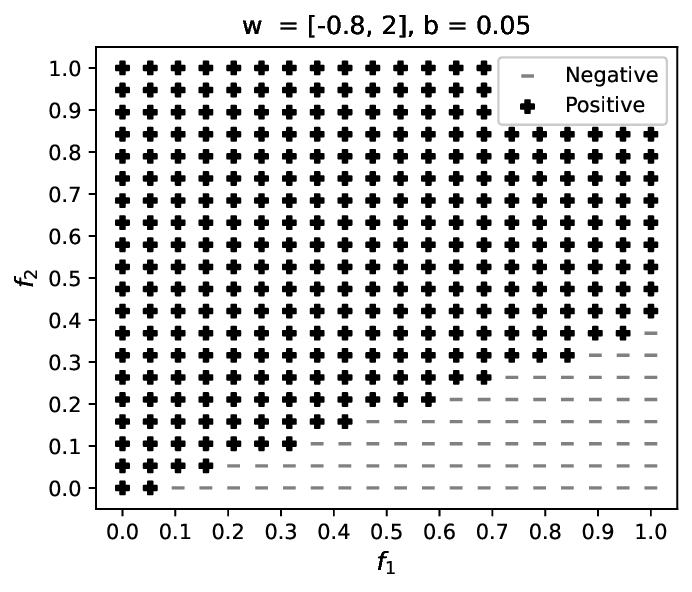}
\caption{\label{fig:Figura 1} Classification through features $f_{1}$ and $f_{2}$.}
\end{figure}

Assume a binary classification problem where $\mathcal{F} = \{f_{1}, f_{2}\}$, and a linear SVC where $w = \{w_1 = -0.8, w_2 = 2\}$ and $b = 0.05$. The features $f_i$ can take values in range $[0, 1]$. A visual representation is depicted in Figure \ref{fig:Figura 1}. Analyzing solely through the values of the weights, it could be assumed that the feature $f_{2}$ is determinant and $f_{1}$ is not, since $|w_2| > |w_1|$. This would mean that, for any instance, feature $f_2$ is more important than feature $f_1$. However, for the instance $\{f_{1}=0.0526, f_2 = 0.3\}$ predicted as class $+1$, feature $f_2$ is not necessary for the prediction, since for any value of $f_{2}$ the class will not change. Therefore, feature $f_1$ is sufficient for the prediction of this instance. Moreover, feature $f_2$ is not sufficient for the prediction of this instance, since for $f_1 = 1.0$ and $f_2 = 0.3$ the prediction would change to $-1$. Therefore, for instance $\{f_{1}=0.0526, f_2 = 0.3\}$, feature $f_{1}$ would be determinant and $f_{2}$ would not.
%Moreover, 
%for any value of $x_{i,1}$ the class would not change.
%\textcolor{red}{[É melhor uma figura como uma malha e nao com os pontinhos]}
%!htpb

While it is trivial to observe if feature $f_{1}$ and $f_{2}$ can be determinant, due to the simplicity of the given decision function, it becomes harder to do so as the number of features increases. Furthermore, the presence of reject option adds another layer of complexity to the problem, turning the interpretation task very non-trivial. Hence, explanations given by weights evaluation may give room for incorrect interpretations. This raises the need for an approach to compute explanations with guarantees of correctness for linear SVCs with reject option.
%a more local approach
%a more rigorous approach.
%instance-based 

%\textcolor{red}{falar que também pode acontecer quando temos uma quantidade de features maior. Também realçar que o caso de reject option não %teria uma maneira fácil de intepretar com base nos pesos das features.}

\section{Explanations For Linear SVMs with Reject Option}\label{Explanations_For_Linear_SVMs_With_Reject_Option}
As depicted before, most methods that generate explanations for ML models are heuristic, which can bring issues about how correct and trustworthy the generated answers are. Added that they tend not to take into account whether reject option is present, these problems are further aggravated. As such, an approach that guarantees the correctness of explanations for ML models with reject option is due.

In this work, we are considering instance-based explanations, i.e. given a trained linear SVC with reject option and an instance $I = \{f_1 = v_1, f_2 = v_2, ..., f_n = v_n\}$ classified as class $c \in \{-1, +1, 0\}$, an explanation $E$ is a subset of the instance $E \subseteq I$ sufficient for the prediction $c$. Therefore, by fixing the values of the features in $E$, the prediction is guaranteed to be $c$, regardless of the values of the other features.

%is based on previous work from the literature \cite{} and it
Our approach is achieved through the encoding of the linear SVC with reject option as a first-order formula, following the ideas of \cite{ignatiev2019abduction} for neural networks. Here we need to take into account the reject option and its encoding as a first-order formula. For example, given a trained linear SVC with reject option defined by $w$, $b$, $t_+$ and $t_-$ as in Equations~\ref{decision_function} and~\ref{reject_predicition_output}, and an instance $I$ classified as class $c = 0$, i.e. in the rejection class, we define the first-order formula $P$:

\begin{equation}\label{P_with_reject}
        \begin{aligned}
             P = (\sum^n_{i=1} w_i f_i  + b \leq t_+) \wedge (\sum^n_{i=1} w_i f_i  + b \geq t_-).
        \end{aligned}    
\end{equation}
Therefore, given an instance $I = \{f_1 = v_1, f_2 = v_2, ..., f_n = v_n\}$ such that its features are defined by $D = \{l_1 \leq f_1 \leq u_1, l_2 \leq f_2 \leq u_2, ..., l_n \leq f_n \leq u_n \}$, an explanation is a subset $E \subseteq I$ such that $E \cup D \models P$. In other words, prediction is guaranteed to be in the rejection class. Instances predicted in other classes are treated in a similar way. For example, if instance $I$ is classified as class $c = +1$, $P$ is defined as $\sum^n_{i=1} w_i f_i  + b > t_+$. It is worth noting that, in these formulas, each $f_i$ is a first-order variable, and $b$, $t_+$, $t_-$ and each $w_i$ are concrete real numbers. 

%\textcolor{red}{Mateus: Sugestão para trocar "concrete real numbers" para $\in {\rm I\!R}$}% É PQ f_i é uma variável/icognita que também está em R, entao estou querendo fazer essa distincao de variável/icognita e constante.

Since succinct explanations seem easier to understand, one would like $E$ to be minimal, that is, for all subset $E' \subseteq E$, $E' \cup D \not\models P$. Therefore, if $E$ is minimal, then each of its subsets does not guarantee the same prediction. Informally, a minimal explanation provides an irreducible explanation that is sufficient for guaranteeing the prediction.

Earlier work \cite{ignatiev2019abduction} outlined Algorithm~\ref{algorithm1} for computing minimal explanations for neural networks. By leveraging the insights from this earlier work, we can effectively employ Algorithm~\ref{algorithm1} to find a minimal explanation for linear SVCs with reject option given $I$, $D$ and $P$. The minimal explanation $E$ of $I$ is calculated by setting $E$ to $I$ and then removing feature by feature from $E$. For example, given a feature $f_i = v_i$ in $E$, if $E \setminus \{f_i = v_i\}, D \models P$, then the value $v_i$ of feature $f_i$ is not necessary to ensure the prediction, and then it is removed from $E$. Otherwise, if $E \setminus \{f_i = v_i\}, D \not\models P$, then
$f_i = v_i$ is kept in $E$ since it is necessary for the prediction. This process is performed for all features as described in Algorithm~\ref{algorithm1}. Then, at the end of Algorithm~\ref{algorithm1}, for the values of features in $E$, the prediction is the same and invariant with respect to the values of the remaining features.

%holds is equivalent to test whether the set of first-order formulas $(E \setminus \{f_i = v_i\}) \cup D \cup \{\neg P\}$ is unsatisfiable, 

By the result in~\ref{entailment_and_unsat}, verifying entailments of the form $(E \setminus \{f_i = v_i\}) \cup D \models P$ can be done by testing whether the set of first-order formulas $(E \setminus \{f_i = v_i\}) \cup D \cup \{\neg P\}$ is unsatisfiable. If $\neg P$ is equivalent to a disjunction $P_1 \vee P_2$ as in \ref{P_with_reject}, then we must check whether $(E \setminus \{f_i = v_i\}) \cup D \cup \{P_1\}$ is unsatisfiable and $(E \setminus \{f_i = v_i\}) \cup D \cup \{P_1\}$ is unsatisfiable, by the result in~\ref{or_unsat}. 

Moreover, since $(E \setminus \{f_i = v_i\}) \cup D$ is a set of linear constraints and $P_1$ and $P_2$ are linear constraints, the unsatisfiability checkings can be achieved by two queries answered by a linear programming (LP) solver. Therefore, if, for example, the set of linear constraints $(E \setminus \{f_i = v_i\}) \cup D \cup \{P_1\}$ has a solution, then $(E \setminus \{f_i = v_i\}) \cup D \not\models P$. From a computational complexity view-point, the linear programming problem is solvable in polynomial time. Then, our approach for computing minimal explanations for linear SVCs with reject option can also be solved in polynomial time. This is achieved by a linear number of calls to an LP solver, which further contributes to the efficiency and feasibility of our approach.

%TODO: on correctness and minimality

%Since $(E \setminus \{f_i = v_i\}) \cup D$ is a set of linear constraints, the unsatisfiability checking can be achieved by a series of queries answered by a linear programming (LP) solver.

%encode linear constraints and $\neg P$ is a disjunction of linear constraints, this unsatisfiability checking can be achieved by a series of queries answered by a linear programming (LP) solver.

%Then, since formulas in $(E \setminus \{f_i = v_i\}) \cup D$ encode linear constraints and $\neg P$ is a disjunction of linear constraints, this unsatisfiability checking can be achieved by a series of queries answered by a linear programming (LP) solver.

%We can use and SMT solver. We choose to use an LP solver since they are optimzed for linear constraints, while SMT solvers can solve more general logical constraints.

%these entailment
%queries are answered by some oracle that solves MILP problems.

%$(E \setminus \{f_i = v_i\}) \cup D \cup \{\varphi \vee \psi  \}$ is unsat iff (E \setminus \{f_i = v_i\}) \cup D \cup \{\varphi\} unsat and (E \setminus \{f_i = v_i\}) \cup D \cup \{\varphi\}

\begin{algorithm}[t]
\caption{Computing a minimal explanation} \label{algorithm1}
\begin{algorithmic}
    \State \textbf{Input:} instance $I$, domain constraints $D$, prediction $P$
    \State \textbf{Output:} minimal explanation $E$
    \State $E \gets I$
    \For{$f_i=v_i \in E$}
        \If{$(E \backslash \{f_i=v_i\}) \cup D \models P$}
            \State $E \gets E \backslash \{f_i=v_i\}$
        \EndIf
    \EndFor
    \State \Return $E$
\end{algorithmic}
\end{algorithm}

\section{Experiments}\label{Experiments}
In this paper, a total of 6 datasets are used. The Vertebral Column and the Sonar datasets are available on the UCI machine learning repository.\footnote{https://archive.ics.uci.edu/ml/datasets.php} The Pima Indians Diabetes dataset is available on Kaggle.\footnote{https://www.kaggle.com/datasets/uciml/pima-indians-diabetes-database} The Iris, the Breast Cancer Wisconsin, and the Wine datasets are available through the scikit-learn package.\footnote{https://github.com/scikit-learn/scikit-learn/tree/main/sklearn/datasets/data} All attributes were scaled to the range $[0, 1]$.

The Iris and the Wine dataset have three classes and have been adapted to be binary classification problems, changed to \textit{setosa-versus-all} and \textit{class\_0-versus-all}, respectively. The other datasets have two classes. The classes were changed to follow values in $\{-1, +1\}$. A summary of the datasets is presented in Table 1.

\begin{table}[htbp]
\caption{Datasets details.}
\label{tab:Datasets Details}
\rowcolors{2}{white}{lightgray} 
\centering % used \centering instead of the center environment since the latter adds additional white space
\begin{tabular}{|c|c|c|c|c|}\hline
Dataset                 & Acronym & $|\mathcal{F}|$ & Negative Patterns & Positive Patterns\\\hline
Iris                    & IRIS & 4 & 50 & 100 \\
Vertebral Column        & VRTC & 6 & 100 & 210 \\
Pima                    & PIMA & 8 & 500 & 268 \\
Wine                    & WINE & 13 & 59 & 119 \\
Breast Cancer Wisconsin & BRCW & 30 & 212 & 357 \\
Sonar                   & SONR & 60 & 111 & 97\\
\hline
\end{tabular}
\end{table}

\textbf{The classifiers}. For each dataset, a linear SVC was trained based on 70\% of the original data, with a regularization parameter $C=1$ and stratified sampling. For finding the rejection region, a value of $w_r = 0.24$ was used together with the decision function outputs based on training data. The selected rejection thresholds were obtained by minimizing the empirical risk, as described in Subsection \ref{Reject_Option_Classification}. All patterns were used for generating explanations, including both the ones used for training and testing. This is due to the focus on explaining rather than model evaluation.

% The entire datasets were used for the experiments, instead of the remaining 30\% unused during the classifier training step, since the objective is to generate explanations instead of model evaluation. \textcolor{red}{No paragrafo anterior vc diz que só usou 70.}

%\noindent For defining the values of $t_+$ and $t_-$ we determine a range of thresholds $T = {(t_{+}^{1},t_{-}^{1}),...,\text{ }(t_{+}^{k},\text{ }t_{-}^{k}) }$, containing the respective possible candidates. The maximum absolute value for $t_+$ in $T$ is the highest output value of the decision function, i.e. the $upper\_limit$, based on the given patterns for generating explanations. Similarly, the maximum absolute value for $t_-$ in $T$ is the lowest output value of the decision function, i.e. the $lower\_limit$,  with respect to the patterns. Thus, the attainable thresholds are achieved through

% \textcolor{red}{A range of thresholds must be defined as possible candidates for the reject region that minimizes the empirical risk. This is accomplished via the lowest and the highest value output from the decision function via the training patterns, that is the lower and upper limit. Thus, the attainable thresholds are achieved through}

For defining the values of $t_+$ and $t_-$, we determine a range of thresholds $T = \{(t_{+}^{1}, t_{-}^{1}), ..., (t_{+}^{k},t_{-}^{k})\}$ containing the respective possible candidates. The maximum absolute value for $t_+$ and $t_-$ in $T$, respectively, is the highest and lowest output value of the decision function, i.e. the $upper\_limit$ and the $lower\_limit$, based on the training patterns. Thus, the attainable thresholds are achieved through
\begin{equation}
%    \begin{cases}
        \begin{aligned}
         T &= \{(i \cdot 0.01 \cdot  upper\_limit,\text{ }i \cdot  0.01 \cdot  lower\_limit) \mid i \in \{1, ..., 100\} \}.
        \end{aligned}    
%    \end{cases}
\end{equation}
Hence, the selected values of $t_+$ and $t_-$ are the pair that minimizes the empirical risk $\hat{R}$. In addition, using $w_r = 0.24$, we obtained the best reject region defined by $t_+$ and $t_-$, the test accuracy of the classifier with RO, and the rejection ratio based on test data. The test accuracy of a classifier with RO is the standard test accuracy applied only to non-rejected instances. Afterward, we determine the number of patterns per class from both training and test data, i.e. the entire datasets. Table \ref{Rejection_Thresholds} details these results and also the test accuracy of the classifier without reject option.

%\textcolor{red}{the accuracy of the classifier (without reject option) and the obtained results.}

\begin{table}[htbp]
\caption{Reject region thresholds using training data. Classifier accuracy without reject option, accuracy with reject option, and rejection ratio for test data. Patterns by class for each entire dataset.}
\label{Rejection_Thresholds}
\rowcolors{2}{white}{lightgray} 
\centering
\resizebox{1.0\textwidth}{!}{
    \begin{tabular}{ |c|c|c|c|c|c|c|c|c|c| }
    \hline
    Dataset & $t_-$ &   $t_+$  & Accuracy w/o RO & Accuracy w/ RO & Rejection & Negative  & Rejected & Positive\\\hline
    IRIS    & -0.0157 & 0.0352  & 100.0\% & 100.00\% & 00.00\% & 50  & 0   & 100 \\\hline
    VRTC    & -0.3334 & 0.8396  & 76.34\% & 89.65\% & 47.92\%  & 22  & 139 & 149\\\hline
    PIMA    & -1.1585 & 0.8312  & 76.62\% & 92.20\% & 59.59\%  & 232 & 474  & 62\\\hline
    WINE    & -0.0259 & 0.0243  & 96.29\% & 96.29\% & 00.56\% & 56  & 1   & 121\\\hline
    BRCW    & -0.4914 & 0.2370  & 97.66\% & 98.76\% & 04.02\%  & 190 & 25  & 354\\\hline
    SONR    & -0.3290 & 0.2039  & 74.60\% & 79.59\% & 20.00\%  & 76  & 43  & 89 \\
    \hline
    \end{tabular}
}
\end{table}

Observe that the reject region for the IRIS dataset did not return any rejected patterns in the dataset. This is likely due to the problem being linearly separable. Therefore, since our experiments rely on explaining the instances present in the dataset, this case of the IRIS dataset is treated as if reject option is not present.

%\textcolor{red}{Falar das acurácias e/ou outra(s) métrica(s) de RO encontradas}
% \textcolor{red}{For the other datasets, the presence of reject option improved the accuracy in all cases. This is expected since the rejected patterns are not taken into account for evaluating accuracy. It is important to note that we were able to achieve $100\%$ accuracy with the WINE dataset while rejecting very few patterns. In addition, there was a substantial increase in accuracy for the VRTC dataset at the cost of rejecting roughly a third of all the test patterns. Different values for $w_r$ can be used to achieve desirable results, depending on the analysis of how impactful a rejection is in each dataset.}
The reject region obtained for the WINE dataset did not reject any of the test patterns, therefore having no change in accuracy, likely due to the fact that the value chosen for $w_r$ penalizes rejections too harshly. For the other datasets, the presence of reject option lead to a higher accuracy in all cases. This is expected since the rejected patterns are not taken into account for evaluation. In addition, there was a substantial increase for both the VRTC and the PIMA dataset at the cost of rejecting roughly 48\% and 60\% of all the test patterns, respectively. Different values for $w_r$ can be used to achieve desirable results, depending on how much a rejection must be penalized in each dataset.
%Use logic-based approach to search for instances in the reject region

\textbf{Anchors}. We compared our approach against the heuristic method Anchors for computing explanations. Anchors was designed to work with traditional multiclass classification problems, i.e. classifiers without reject option. Then, since we are explaining predictions of a linear SVC with reject option, we used the Anchors explanation algorithm to treat the classifier with reject option as if it were a traditional classifier with classes in $\{-1, 0, +1\}$.

%Due to the characteristics of the datasets, a tabular anchors explainer was used. It was set with the [$-1, 0, 1$] classes, enabling explanations for the rejected class.

\textbf{Our approach}. The prototype implementation of our approach\footnote{https://github.com/franciscomateus0119/Logic-based-Explanations-for-Linear-Support-Vector-Classifiers-with-Reject-Option} is written in Python and follows Algorithm~\ref{algorithm1}. As an LP solver to check unsatisfiability of sets of first-order sentences, we used Coin-or Branch-and-Cut (CBC)\footnote{https://github.com/coin-or/Cbc}. Moreover, the solver is accessed via the Python API PuLP\footnote{https://github.com/coin-or/pulp}.

%The Coin-or Branch-and-Cut (CBC) solver was used for generating explanations and for validating Anchors outputs.} 
%The Pulp\footnote{https://github.com/coin-or/pulp} LP modeler package provided the tools for the problem modeling.

%\subsection{Linear Programming Problem Modeling}\label{Linear_Programming_Problem_Modeling}
%\textcolor{red}{For the LP problem modeling, four types of information are needed: the patterns, the predicted labels, the classifier decision function, and the reject region thresholds. }

\subsection{Results}\label{Results}

%No rejected results for the IRIS dataset, likely due to the problem being linearly separable. (removed from table)

A per-instance explanation is done for each dataset. Both mean elapsed time (in seconds) and the mean of the number of features in explanations are used as the basis for comparison. The results based on the former and the latter are presented in Table \ref{tab:Time_comparison} and Table \ref{tab:Size_comparison}, respectively. Following the IRIS dataset description in Table \ref{Rejection_Thresholds}, there are no results for the rejected class.
\begin{table}[htbp]
\caption{Time comparison between our approach and Anchors (seconds). Less is better.}
\label{tab:Time_comparison}
%\rowcolors{3}{lightgray}{white} 
\centering
\resizebox{1.0\textwidth}{!}{
    \begin{tabular}{ |c|c|c|c|c|c|c| }
    \hline
    \multicolumn{1}{|c|}{\multirow{2}{*}{Dataset}} & \multicolumn{2}{c|}{Negative} & \multicolumn{2}{c|}{Rejected} & \multicolumn{2}{c|}{Positive}\\
    \cline{2-7}
                        & {Anchors} & {Ours}  & {Anchors} & {Ours}       & {Anchors} & {Ours}\\ \hline
\rowcolor{lightgray}    IRIS & 0.13  $\pm $ 0.045  & 0.04  $\pm $ 0.004   & -      & -               & 0.05  $\pm $  0.024 & 0.04  $\pm $ 0.005\\\hline
\rowcolor{white}        VRTC & 0.27  $\pm $ 0.049  & 0.07  $\pm $ 0.004   & 0.33  $\pm $ 0.147   & 0.12  $\pm $ 0.018 & 0.28  $\pm $ 0.156 & 0.07  $\pm $ 0.002\\\hline
\rowcolor{lightgray}    PIMA & 0.85  $\pm $ 0.162  & 0.10  $\pm $ 0.001   & 0.27  $\pm $ 0.302   & 0.16  $\pm $ 0.021 & 0.61  $\pm $ 0.108 & 0.10  $\pm $   0.002\\\hline
\rowcolor{white}        WINE & 3.01  $\pm $ 0.335  & 0.17  $\pm $ 0.002   & 1.60  $\pm $ 0.000   & 0.21  $\pm $ 0.000 & 0.38  $\pm $ 0.532 & 0.17  $\pm $ 0.002\\\hline
\rowcolor{lightgray}    BRCW & 26.58  $\pm $ 4.268 & 0.42  $\pm $ 0.015   & 20.66  $\pm $ 3.506  & 0.55  $\pm $ 0.072 & 5.84  $\pm $ 9.686 & 0.41  $\pm $ 0.013\\\hline
\rowcolor{white}        SONR & 233.26  $\pm $ 70.550  & 0.86  $\pm $ 0.011 & 257.51  $\pm $ 16.267 & 1.27  $\pm $ 0.258 & 246.14  $\pm $ 51.82 & 0.86  $\pm $  0.019\\
    \hline
    \end{tabular} 
}
\end{table}

\begin{table}[htbp]
\caption{Explanation size comparison between our approach and Anchors. Less is better.}
\label{tab:Size_comparison}
\centering
\resizebox{1.0\textwidth}{!}{
    \begin{tabular}{ |c|c|c|c|c|c|c| }
    \hline
    \multicolumn{1}{|c|}{\multirow{2}{*}{Dataset}} & \multicolumn{2}{c|}{Negative} & \multicolumn{2}{c|}{Rejected} & \multicolumn{2}{c|}{Positive}\\
    \cline{2-7}
                        & {Anchors} & {Ours}  & {Anchors} & {Ours} & {Anchors} & {Ours}\\ \hline
\rowcolor{lightgray}    IRIS & 3.52  $\pm $ 0.854  & 3.00  $\pm $ 0.000   & -  & -& 1.89  $\pm $ 0.733 & 2.2  $\pm $ 0.400\\ \hline
\rowcolor{white}        VRTC & 4.95  $\pm $  1.580  & 5.18  $\pm $  0.574   & 4.56  $\pm $ 1.941 & 4.99  $\pm $ 0.515 & 4.24  $\pm $  2.077 & 4.20  $\pm $ 1.118\\ \hline
\rowcolor{lightgray}    PIMA & 7.50  $\pm $ 1.534  & 5.13  $\pm $ 0.944   & 3.02  $\pm $  2.249 & 4.71  $\pm $ 0.897 & 6.37  $\pm $ 2.541 & 4.85  $\pm $ 1.119\\ \hline
\rowcolor{white}        WINE & 12.98  $\pm $ 0.132 & 7.44  $\pm $ 1.252   & 13.00  $\pm $ 0.000 & 12.00  $\pm $ 0.000 & 2.80  $\pm $ 2.803 & 7.04  $\pm $ 1.605\\ \hline
\rowcolor{lightgray}    BRCW & 29.17  $\pm $ 4.372  & 22.29  $\pm $ 3.739 & 29.64  $\pm $ 1.763  & 27.08  $\pm $ 1.016 & 7.70  $\pm $ 11.024 & 23.08  $\pm $ 2.208\\ \hline
\rowcolor{white}        SONR & 51.10  $\pm $ 20.075  & 50.39  $\pm $ 4.283 & 57.95  $\pm $ 9.233  & 56.16  $\pm $ 1.413 & 54.86  $\pm $ 14.625 & 51.78  $\pm $ 1.413\\ 
    \hline
    \end{tabular}
}
\end{table}

%\textcolor{red}{Realçar o motivo das células em branco no IRIS. Além disso, comentar sobre alguma instância para a qual sua explicação é similar %ao que é descrito na secao 3}

Our approach has shown to be up to, surprisingly, roughly $286$ times faster than Anchors. This happened for the SONR dataset, where the mean elapsed time of our method is $0.86$ seconds against $246.14$ seconds for Anchors. These results show how much harder it can be for Anchors to give an explanation as the number of features increases. Furthermore, it is worth noting that our approach needs much less time to explain positive, negative, and rejected classes overall. 

%it can also be more succinct, outputting up to 43\% smaller explanations in other cases.
While our approach obtained more extensive explanations in certain cases, 
it also demonstrated the ability to be more succinct, generating explanations that were up to $43\%$ smaller in other cases. In addition, our method is able to maintain the correctness of explanations. This highlights an expressive advantage of our approach over the heuristic nature of Anchors. While Anchors may provide explanations with fewer features in some cases, this is achieved due to the lack of formal guarantees on correctness.

%This demonstrates an expressive advantage against the heuristic nature of anchors, which may output answers with fewer features due to having no formal guarantees of correctness.

It is important to note that cases similar to what we discussed in Section~\ref{Linear_SVCs_May_Not_Be_Locally_Interpretable} occurred for the IRIS dataset. Through weights $\mathbf{w} = \{w_1 = 0.8664049, w_2 = -1.42027753, w_3 = 2.18870793, w_4 = 1.7984087\}$ and $b = -0.77064659$, obtained from the trained SVC, it could be assumed that feature $f_1$ might possibly be not determinant for the classification. Explanation results for both negative and positive classes are presented in Table \ref{tab:Iris_Important_Features}, depicting the number of times each feature was present in explanations with our approach.%considered determinant with our approach

\begin{table}[H]%htbp
\caption{Number of times each feature is determinant for the IRIS dataset classification.}
\label{tab:Iris_Important_Features}
\rowcolors{2}{white}{lightgray} 
\centering
\resizebox{0.4\textwidth}{!}{
    \begin{tabular}{ |c|c|c|c|c|c| }
    \hline
    Class & $f_1$& $f_2$& $f_3$& $f_4$& Patterns\\\hline
    \cline{1-5}
    Positive          & 0  & 28 & 100 & 92  & 100\\ \hline
    Negative          & 1  & 49 & 50  & 50  & 50\\ 
    \hline
    \end{tabular}
}
\end{table}

%\noindent While the analysis of the weights for the positive patterns does show that feature $f_1$ is not determinant in any case, the same did not happen for the negative ones. Our approach found an instance where $f_1$ is a determinant feature. If $\mathbf{f_i} =\{f_1 = 0.05555556,f_2,f_3 = 0.05084746,f_4 = 0.08333333\}$, then for any value that $f_2$ assumes the predicted class will not change. Furthermore, a similar case This is substantial for our approach since it demonstrates its capability of finding such cases.

For example, for the instance $\{f_1 = 0.05555556, f_2 = 0.05833333,f_3 = 0.05084746,f_4 = 0.08333333\}$ in class $-1$, feature $f_1$ was in the explanation, while feature $f_2$ was not present. Then, for any value that $f_2$ assumes, the predicted class will not change. Furthermore, similar cases happened where $f_2$ was in the explanation while $f_4$ was not, even though $|w_4| > |w_2|$.

%Our approach found an instance, for the negative class, where $f_1$ is a determinant feature. If $\mathbf{f_i} =\{f_1 = 0.05555556,f_2,f_3 = 0.05084746,f_4 = 0.08333333\}$, 

Similar occurrences are present in other datasets, reinforcing that the points discussed in Section~\ref{Linear_SVCs_May_Not_Be_Locally_Interpretable} are not uncommon. As one more illustrate example, consider the linear SVC with reject option trained on the VRTC dataset such that $\mathbf{w} = \{w_1 = 0.72863148,$ $w_2 = 1.97781269,$ $w_3 = 0.85680605,$ $w_4 = -0.32466632, $ $w_5 = -3.42937211,$ $w_6 = 2.43522629\}$, $b = 1.10008469$, $t_- = -0.3334$ and $t_+ = 0.8396$. For the instance $\{f_1 = 0.25125386,f_2 = 0.4244373,f_3 =0.7214483, f_4 = 0.20007403, f_5 = 0.71932466,f_6 = 0.15363128\}$ in the reject class, feature $f_3$ was not in the explanation, while $f_4$ is present in the explanation, despite the fact that $|w_3| > |w_4|$. This is substantial for our approach since it demonstrates its capability of finding such cases.

%for the rejected class in the VRTC dataset through $\mathbf{w} = \{w_1 = 0.72863148, w_2 = 1.97781269, w_3 = 0.85680605, w_4 = -0.32466632, w_5 = -3.42937211, w_6 = 2.43522629\}$ and $b = 1.10008469$, if 

%$\mathbf{f_i} =\{f_1 = 0.25125386,f_2 = 0.4244373,f_3,f_4 = 0.20007403, f_5 = 0.71932466,f_6 = 0.15363128\}$ then for any value that $f_3$ assumes the predicted class will not change, where $f_4$ is a determinant feature and $|w_3| > |w_4|$.} This is substantial for our approach since it demonstrates its capability of finding such cases.

\section{Conclusions}\label{Conclusions}
In this paper, we propose a logic-based approach to generate minimal explanations for a classifier with reject option while guaranteeing correctness. A trained linear SVC with reject option is used as a target model for explainability. Our approach is rooted in earlier work on computing minimal explanations for standard machine learning models without reject option. Therefore, we encode the task of computing explanations for linear SVCs with reject option as a logical entailment problem. Moreover, we use an LP solver for checking the entailment, since all first-order sentences are linear constraints with real variables and at most one disjunction occurs.

%builds on work from the literature of logic-based explanations for models without reject option.

%it is also able to 
Our method is compared against Anchors, one of the most well-known heuristic methods, through six different datasets. We found that not only the proposed method takes considerably less time than Anchors, but also reduces the size of explanations for many instances. Our approach achieved astonishing results in terms of efficiency, surpassing Anchors by an impressive factor of up to approximately 286 times.

%Logic based for NN and RF
Our approach can be further improved in future work. For example, it can be easily adapted to other classifiers with reject option, such as neural networks and random forests. In addition, it can be adjusted for a non-linear SVM. Another improvement is the generalization of explanations by allowing a range of values for each feature, rather than solely considering equality. This improvement may enable a more comprehensive understanding of the model.

%better understanding of which interval of values the features are actually relevant for the classification, contributing to a more comprehensive understanding of the model.

%instead of only equality. 
%Another improvement is adding value ranges in the explanations, allowing for a better understanding of which value intervals the features are actually relevant for the classification.

%, such as the polynomial and the radial-basis function (RBF), which is a function that transforms the input data space to a higher-dimensional space where it is more manageable to deal with non-linear problems.

%by \textcolor{red}{being adapted to other classifiers that exist in the literature, such as neural networks, decision trees, random forests, and boosted trees. In addition, it can be adjusted for a non-linear SVM kernel, such as the polynomial and the radial-basis function (RBF), which is a function that transforms the input data space to a higher-dimensional space where it is more manageable to deal with non-linear problems. }

\subsubsection{Acknowledgments.} The authors thank FUNCAP and CNPq for partially supporting our research work.

\bibliographystyle{splncs04}
\bibliography{refs.bib}
%





% LNCS Homepage, http://www.springer.com/lncs}. Last accessed 4
% \end{thebibliography}
\end{document}